\documentclass[11pt,a4paper]{article}
\usepackage[hyperref]{conll-2019}
\usepackage{times}
\usepackage{latexsym}
\usepackage{makecell}
\usepackage{multirow}
\usepackage{url}
\usepackage{booktabs}       
\usepackage{amsfonts}       
\usepackage{nicefrac}       
\usepackage{microtype}      
\usepackage{amsmath}
\usepackage{bbm}
\usepackage{color}
\usepackage{multirow}
\usepackage{graphicx}
\usepackage{caption}
\usepackage{subcaption}
\usepackage{amssymb}
\usepackage{enumitem}
\usepackage{makecell}
\usepackage{float}
\usepackage{mwe,tikz}
\usepackage[percent]{overpic}

\usepackage{linguex}

\aclfinalcopy 

\usepackage{pifont}
\newcommand{\cmark}{\ding{51}}%

\newcommand{\lmms}{\ensuremath{\mathcal{L}_\text{MMS}}}
\newcommand{\ltrip}{\ensuremath{\mathcal{L}_\text{T}}}

\makeatletter
\newcommand*{\centerfloat}{%
  \parindent \z@
  \leftskip \z@ \@plus 1fil \@minus \textwidth
  \rightskip\leftskip
  \parfillskip \z@skip}
\makeatother

\title{Large-scale representation learning from visually grounded untranscribed speech}

\author{Gabriel Ilharco$\dagger$\thanks{$\ $ Work done as a member of the Google AI Residency Program.} \quad Yuan Zhang$^{\ddagger}$ \quad Jason Baldridge$^{\ddagger}$\\
  $^\dagger$Paul G. Allen School of Computer Science \& Engineering,\\
  University of Washington,
  Seattle, WA, USA \\
  $^\ddagger$Google Research, Mountain View, CA, USA \\
  {\tt gamaga@cs.washington.edu}, {\tt \{zhangyua,jridge\}@google.com} \\}

\date{}

\begin{document}
\maketitle
\begin{abstract}

Systems that can associate images with their spoken audio captions are an important step towards visually grounded language learning. We describe a scalable method to automatically generate diverse audio for image captioning datasets. This supports pretraining deep networks for encoding both audio and images, which we do via a dual encoder that learns to align latent representations from both modalities. We show that a masked margin softmax loss for such models is superior to the standard triplet loss. We fine-tune these models on the Flickr8k Audio Captions Corpus and obtain state-of-the-art results---improving recall in the top 10 from 29.6\% to 49.5\%. We also obtain human ratings on retrieval outputs to better assess the impact of incidentally matching image-caption pairs that were not associated in the data, finding that automatic evaluation substantially underestimates the quality of the retrieved results.

\end{abstract}

\section{Introduction}

Natural language learning in people starts with speech, not text. Text is tidy: it comes in convenient symbolic units that vary little from one writer to another. Speech is continuous and messy: the sounds used to convey a given word are modified by those of surrounding words, and the rate of speech, its pitch, and more vary across speakers and even for the same speaker in different contexts. As such, problems involving speech provide distinct challenges and opportunities for learning language representations that text-based work---which represents the vast majority---gets a free pass on.

\begin{figure}
  \centering   
  \includegraphics[clip, width=\linewidth]{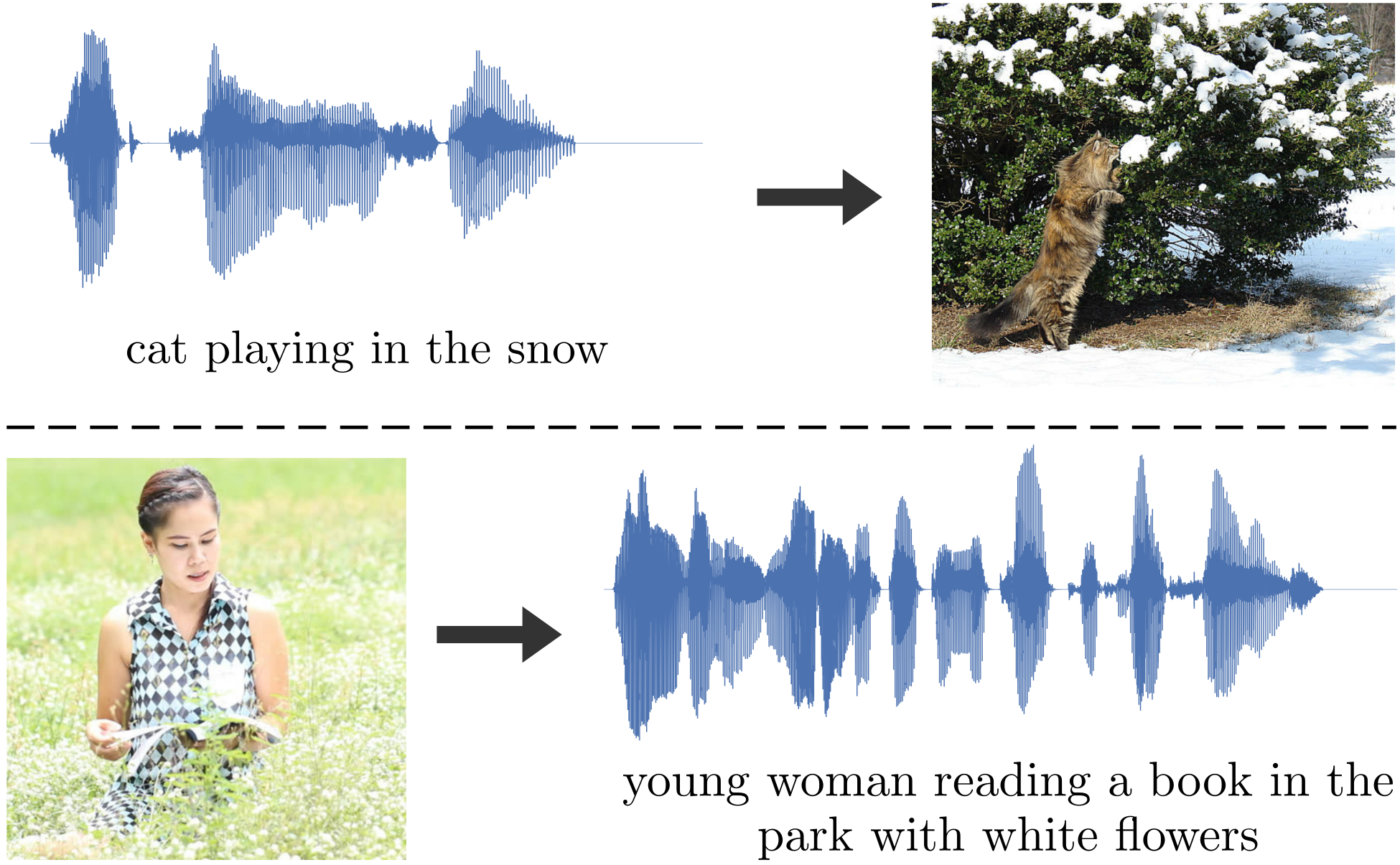}
  \caption{Models that encode speech segments and images into a shared latent space enable images to be retrieved using their audio descriptions (top) and to associate images with spoken captions (bottom). Text captions are provided for clarity; only speech and images are used by the models.}
\label{fig:intro_summary}
\end{figure}

Recent work has explored various means to transform raw speech into symbolic forms with little or no supervision \cite{park2007unsupervised, varadarajan2008unsupervised, ondel2016variational, kamper2017segmental, bhati2018phoneme}. However, learning natural language starts with grounded, contextualized speech. While infants as young as 8-months-old can segment word-like units without non-linguistic information \cite{juscyzk:aslin:1995} and adults can learn to segment words in artificial languages \cite{Saffran1996WordST}, a learner must ultimately ground their representations of linguistic sequences \cite{harnad1990symbol} to effectively use them to refer to objects, events and more. Furthermore, learning from rich perceptual data and interactions can be more efficient as it provides additional cues to the identities of words and their meaning in context. 
 
We address the problem of relating images to audio captions that describe them (Figure \ref{fig:intro_summary}), building on 
previous research into learning from visually grounded, untranscribed speech \cite{harwath2015deep, sun2016look, harwath2016unsupervised, chrupala2017representations, kamper2017visually, chrupala2018symbolic, harwath2019towards}. Such problem settings provide opportunities both to improve our theoretical understanding of language as well as to realize gains on practical problems---including voice interaction with virtual assistants, image retrieval based on speech, and generally better supporting people with visual impairments.

Our contribution is to improve performance on bidirectional speech/image retrieval through better data and better models for learning fixed dimensional latent representations of both modalities. We construct a synthetic speech caption dataset for pretraining by applying text-to-speech (TTS) on Conceptual Captions \cite{sharma2018conceptual}, a dataset with 3.3 million diverse image-caption pairs. Unlike \citet{chrupala2017representations}, who similarly applied TTS to MS-COCO \cite{chen2015microsoft}, we inject diversity by varying the voice, speech rate, pitch and volume gain on every synthetically produced audio caption.  We refer to the resulting dataset as Conceptual Spoken Captions (CSC). CSC's scale allows us to train deeper models than previous work. We use Inception-ResNet-v2 \cite{szegedy2017inception} to encode both the audio and visual modalities in a dual encoder model, pretraining on CSC and then fine-tuning and evaluating on human speech in the smaller Flickr Audio Caption Corpus (FACC) \cite{harwath2015deep}. Using an adapted batch loss function rather than the triplet loss used in previous work, we substantially improve on the previous state-of-the-art for the standard FACC retrieval tasks.

Image captioning datasets contain positively paired items---but that does not imply that a random image and caption cannot also be a valid match. For instance, in FACC there are many spoken captions about beaches and sunsets and plenty of images that match these captions; two different images with descriptions \textit{``A surfer is riding a wave."} and \textit{``A man surfs the wave"} are likely compatible. It is of course not feasible to exhaustively annotate all pairwise associations, so we have human raters judge the top five retrieved results for two models to assess the impact of this aspect of the data on automatic retrieval metrics used thus far. Unsurprisingly, models retrieve many compatible results that are unpaired in FACC: with the human evaluations, we find consistent increases in recall.

\section{Data}

Larger training datasets support better performance and generalization \cite{banko2001scaling, halevy2009unreasonable, sun2017revisiting}, especially for deep models. Collecting labels from people has become easier via crowd computing \cite{buhrmester2011amazon}, but is still expensive and remains a bottleneck for creating broad and representative datasets. This motivates the case for exploiting incidental annotation \cite{roth-incidental} and automating some aspects of dataset creation. The current trend of using machine translation systems to produce augmented datasets for machine translation itself \cite{sennrich-etal-2016-improving} and for monolingual tasks like classification \cite{wei2018fast} and paraphrasing \cite{wieting-gimpel-2018-paranmt} is a good example of this. 

For speech image captioning, \citet{chrupala2017representations} used a Text-to-Speech (TTS) system to create audio from the textual captions given in the MS-COCO dataset, resulting in 300k unique images with 5 spoken captions each.  We scale this idea to the larger and more diverse textual Conceptual Captions dataset with 3.3 million unique image and captions, additionally modifying the produced speech by using multiple voices and random perturbations to the rate, pitch and audio. Our goal is to make the resulting data more effective for pretraining models so they can learn more efficiently on smaller amounts of human speech.

\subsection{Conceptual Captions}
\label{sec:cc}

Image captioning datasets have ignited a great deal of research at the intersection of the computer vision and natural language processing communities \cite{lin2014microsoft, vinyals2015show, bernardi2016automatic, anderson2018bottom}. Getting annotators to provide captions works well with crowd computing, but \citet{sharma2018conceptual} exploit incidental supervision for this task to obtain greater scale with their Conceptual Captions dataset. It contains 3.3 million pairs of image and textual captions, where pairs are extracted from HTML web pages using the \textit{alt-text} field of images as a starting point for their descriptions.

The textual captions are processed in a hypernymization stage. Named entities and syntactic dependency annotations are obtained using Google Cloud Natural Language APIs, which are matched to hypernym terms using the Google Knowledge Graph Search API. Proper nouns, numbers, units, dates, durations and locations are removed; identified named-entities are substituted with their hypernym, merging together analogous terms when possible. For example, the original alt-text \ref{ex:alt-text} is converted to the conceptual caption \ref{ex:cc}.

\ex. \label{ex:alt-text} \textbf{alt-text}: \textit{Musician Justin Timberlake performs at the 2017 Pilgrimage Music \& Cultural Festival on September 23, 2017 in Franklin, Tennessee.}

\ex. \label{ex:cc} \textbf{conceptual caption}: \textit{pop artist performs at the festival in a city.}

There are many sequential filtering steps for improving the quality of the captions---see \citet{sharma2018conceptual} for a thorough description. As quality control, a random sample of 4K conceptual captions were rated by human annotators, and 90.3\% were judged ``good" by at least 2 out of 3 raters. 

\subsection{Conceptual Spoken Captions}

We use TTS to generate a high-fidelity spoken sentence for each of the 3.3 million textual captions in the Conceptual Captions dataset.\footnote{The alt-text does not come with the dataset and cannot be redistributed, so we focus on the conceptual captions for ease of experimentation and reproducibility.} We use the Google Cloud Speech API\footnote{\href{https://cloud.google.com/text-to-speech/}{https://cloud.google.com/text-to-speech/}} for TTS. Internally, the service uses a WaveNet model \cite{van2016wavenet} to generate audio. For diversity, the speech is synthesized using parameter variations, as follows:

\begin{itemize}
    \item \textit{Voice}, which is sampled uniformly from a set of 6 different voices generated using a WaveNet model for American English.
    \item \textit{Speaking rate} controls the speed of the synthesized audio. A speaking rate of 1.0 means the normal speed of a given voice, while a speaking rate of 2.0 means twice as fast. When synthesizing the data, we draw this parameter from a Gaussian distribution $\sim \mathcal{N}(1.0, 0.1^2)$.
    \item \textit{Pitch} controls how high/deep the voice is. For example, if set to 1, this means the voice will be synthesized 1 semitones above the original pitch. This parameter is drawn from a Gaussian distribution $\sim \mathcal{N}(0.0, 1.0^2)$.
    \item \textit{Volume gain} controls a gain in dB with respect to the normal native signal amplitude. If set to 0, the voice is synthesized without alterations in volume. This parameter is drawn from a Gaussian distribution $\sim \mathcal{N}(0.0, 2.0^2)$.
\end{itemize}

To avoid degenerate cases, we clip the values sampled from the Gaussian distributions described above such that they are never more than 2 times the standard deviation from the mean. All spoken captions are generated in 16000 Hz. 

\subsection{Flickr Audio Caption Corpus}

The Flickr Audio Caption Corpus (FACC) \cite{harwath2015deep} consists of 40,000 pairs of images and spoken captions, with 8000 unique images, of which 1000 are held for validation and 1000 for testing. The spoken captions are generated from humans reading the textual captions from the Flickr8k dataset \cite{hodosh2013framing}, originally crowd-sourced and based on images from Flickr. 

We use FACC for evaluation, both when pretraining on Conceptual Spoken Captions and when training on FACC from scratch. Like previous work, the core evaluation considered is retrieval of the known paired image given an audio caption within some top-k set of retrieved items (e.g. R@1 for whether the first item retrieved is the paired one and R@10 for whether it is in the top ten results). We also conduct human evaluations on retrieval outputs to detect the presence of unpaired but matching image-caption pairs identified by the models and thereby better assess their impact on performance.
\begin{figure}
    \centering
    \includegraphics[width=0.75\linewidth]{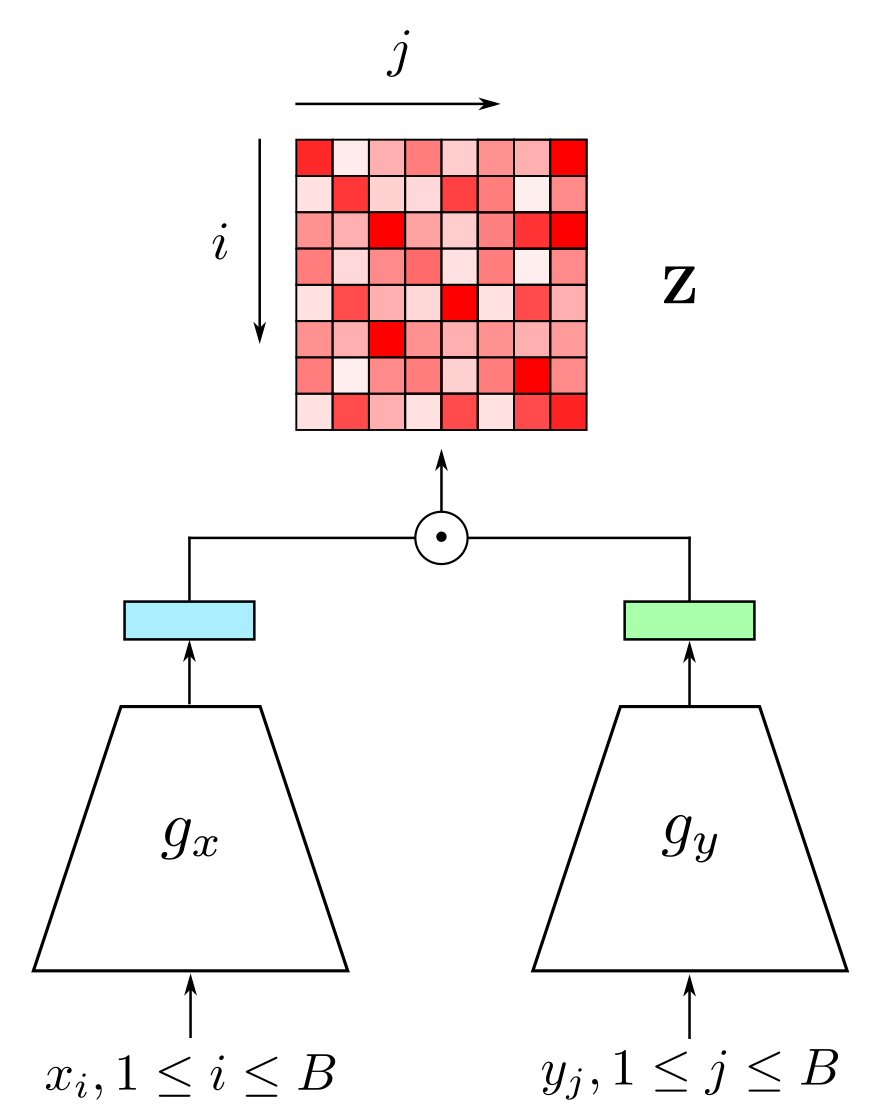}
    \caption{Dual-encoder model architecture.}
    \label{fig:model}
\end{figure}

\section{Model}

Dual encoders are used in a wide range of applications, including signature verification  \cite{bromley1994signature}, object tracking \cite{bertinetto2016fully}, sentence similarity \cite{mueller2016siamese}, improving neural machine translation \cite{yang2019improving} and many others. The core of this set of architectures is a simple two-tower model illustrated in Figure \ref{fig:model}, where inputs $x \in \mathcal{X}$ are processed by an encoder $g_x$ and inputs $y \in \mathcal{Y}$ by a second encoder $g_y$. The inputs may come from the same distribution---or they may be from entirely different sources or modalities. The towers may share the same architecture and weights---or they can be completely unlike and disconnected.

These models are standard in audiovisual image captioning \citep{harwath2015deep,chrupala2018symbolic,harwath2018jointly}. In this setting, the dual encoder model, is composed by a visual tower, $g_{vis}$, processing the images, and an audio tower, $g_{aud}$, processing the spoken captions. The model is trained to map both modalities into a joint latent space. Here, we extend previous work to consider a batched margin loss, which we show to be superior for learning dense representations for retrieval.

\paragraph{Notation.} The inputs are processed in batches of size $B$. For each input $x_k$ and $y_k$ in the batch, $1 \le k \le B$ , let $g_x(x_k)$ and $g_y(y_k)$ be their latent representations extracted by the corresponding tower.  We define the $B\times B$ matrix $\mathbf{Z}$ as the similarity between the latent representations for each pair of elements in the batch. A natural choice for that similarity is the dot product between the latent representations: 
\begin{equation}
    \mathbf{Z}_{ij} = g_x(x_i) \cdot g_y(y_j)
    \label{eq:Z}
\end{equation}
As shown in Figure \ref{fig:model}, $\mathbf{Z}$ encodes all pairwise associations in the batch. However, an additional aspect of some datasets must be taken into account: often times the same input $x$ can match multiple inputs $y$ or vice-versa---for instance, both Flickr8k and MS-COCO have multiple captions for the each image. To respect these pairs when they land in the same batch---and thus not penalize models for (correctly) associating them---we define a $B\times B$ masking matrix $\mathbf{M}$: 
\begin{equation}
\mathbf{M}_{ij} = 
\begin{cases}
 0, & \text{if $x_i$ matches $y_j$} \\
 1, & \text{otherwise}
\end{cases}
\label{eq:M}
\end{equation}
\noindent
All pairs $(x_k, y_k)$ match and this equivalence is transitive, so $\mathbf{M}$ is symmetric and all diagonal elements $\mathbf{M}_{kk}$, $1 \le k \le B$ are zero.

\paragraph{Triplet Loss.} Both \citet{chrupala2018symbolic} and \citet{harwath2018jointly} (and their previous work) employ the triplet loss function given in Equation \ref{eq:harwath_loss}.
\begin{align}
\begin{split}
    \mathcal{L_{\text{T}}} = \sum_{k = 1}^B\big(&\max(0, \mathbf{Z}_{km} - \mathbf{Z}_{kk} + \delta) +\\ &\max(0, \mathbf{Z}_{nk} - \mathbf{Z}_{kk} + \delta)\big)
    \label{eq:harwath_loss}
\end{split}
\end{align}
\noindent
For each value $k$, $m$ is randomly drawn from a uniform distribution over indices $j$ such that $M_{kj} = 1$, and $n$ over indices $i$ such that $M_{ik} = 1$.

\begin{table*}
\setlength\tabcolsep{2.8pt}
    \centering
    \begin{tabular}{lccccccccccc}
    & & \multicolumn{5}{c}{\textbf{Speech to Image}} & \multicolumn{5}{c}{\textbf{Image to Speech}} \\\cmidrule(lr{0.1cm}){3-7}\cmidrule(l{0.1cm}){8-12}
    Loss & Batch Size & R@1 & R@5 & R@10 & R@50 & R@100  & R@1 & R@5 & R@10 & R@50 & R@100\\\hline
    $\mathcal{L}_{\text{T}}$ & 48 & .037 & .109 & .165 & .367 & .474 & .031 & .101 & .155 & .346 & .455
    \\\Xhline{.5\arrayrulewidth}
    \multirow{3}{*}{$\mathcal{L}_{\text{MMS}}$}
    & 12 & .025 & .083 & .129 & .311 & .432 & .024 & .083 & .132 & .315 & .433 \\
    & 24  & .054 & .143 & .206 & .418 & .533 & .046 & .137 & .197 & .411 & .520\\
    & 48 & \textbf{.078} & \textbf{.204} & \textbf{.282} & \textbf{.499} & \textbf{.604} & \textbf{.074} & \textbf{.194} & \textbf{.269} & \textbf{.485} & \textbf{.587}
    \end{tabular}
    \caption{Performance on the validation set of Conceptual Spoken Captions, comparing different loss functions and batch sizes.}
    \label{tab:results_cac}
\end{table*}

\paragraph{Masked Margin Softmax Loss.} The triplet loss (\ref{eq:harwath_loss}) used previously misses opportunities to learn against a wider set of negative examples, namely all those in the batch that are not known to be positively associated (i.e., $\mathbf{M}_{ij}=1$). To exploit these additional negatives, we minimize the Masked Margin Softmax (MMS) loss function, inspired by \citet{henderson2017efficient} and \citet{yang2019improving}.  MMS simulates $x$-to-$y$ and $y$-to-$x$ retrievals inside the batch. It is defined at a high level as:
\begin{equation}
    \mathcal{L_{\text{MMS}}} = \mathcal{L}_{xy} + \mathcal{L}_{yx}
    \label{eq:loss}
\end{equation}
\noindent
$\mathcal{L_{\text{MMS}}}$ is the sum of losses defined over $x$-to-$y$ (Eq. \ref{eq:l1}) and $y$-to-$x$ (Eq. \ref{eq:l2}) in-batch retrievals.
\begin{equation}
    \mathcal{L}_{xy} = - \dfrac{1}{B} \sum_{i = 1}^B \log \dfrac{e^{\mathbf{Z}_{ii} - \delta}}{e^{\mathbf{Z}_{ii} - \delta} + \sum_{j = 1}^B\mathbf{M}_{ij} e^{\mathbf{Z}_{ij}}}
    \label{eq:l1}
\end{equation}
\begin{equation}
    \mathcal{L}_{yx} = - \dfrac{1}{B} \sum_{j = 1}^B \log \dfrac{e^{\mathbf{Z}_{jj} - \delta}}{e^{\mathbf{Z}_{jj} - \delta} + \sum_{i = 1}^B\mathbf{M}_{ij} e^{\mathbf{Z}_{ij}}}
    \label{eq:l2}
\end{equation}
\noindent
These are equivalent to a cross-entropy loss after a column-wise or row-wise softmax on the matrix $\mathbf{Z}$, subject to the masking constraints in $\mathbf{M}$ and margin $\delta$.

The margin hyperparameter $\delta$ is gradually increased as training progresses. Empirically, we found that, with a fixed $\delta$, large values lead to unstable performance in early training, while small values lead to negligible results in final performance. Starting with a small $\delta$ and increasing it does not hurt early training and forces the model to learn from a harder task later on. There many ways to increase $\delta$ along training---e.g. linearly, quadratically, and exponentially. The latter is used in this work.

Contrasting Equations \ref{eq:harwath_loss} and \ref{eq:loss}, the former chooses a negative sample randomly, while the latter takes advantage of all negative pairs in the batch and thus improves sample efficiency. $\mathcal{L}_\text{MMS}$ has three main differences with \citet{yang2019improving}: (1) a masking term that accounts for the fact that there might be multiple positive choices in the batch for a given input; (2) a varying margin term $\delta$, which is  increased during training; (3) a log term that makes MMS more closely resemble a cross-entropy loss.
\section{Experiments}
\label{sec:experiments}

\subsection{Experimental settings}
\label{sec:settings}

\paragraph{Image preprocessing.} During training, data augmentation is performed by randomly distorting the brightness and saturation of images. Each image is also randomly cropped or padded such that at least 67\% of the area of the original image is covered, and re-scaled if necessary to 299$\times$299. During evaluation, we do not perform color distortions, and we crop/pad the central portion of the images.

\paragraph{Audio preprocessing.} We extract 128 Mel-Frequency Cepstral Coefficients (MFCCs) from the raw audio signals using a window size of 20ms. The audio signals have a sampling rate of 16000Hz. We compute features every 10ms, such that each window has a 50\% overlap with its neighbors. During training, we randomly crop/pad the MFCCs in the temporal dimension, and perform data augmentation as in \citet{park2019specaugment}, using one mask with a frequency mask parameter of 20 and a time mask parameter of $40$. We do not perform time warping.

\paragraph{Encoders.} Both audio and image encoders are Inception-ResNet-v2 networks \cite{szegedy2017inception}, allowing the model to reap the benefits of relatively low computational cost, fast training and and strong performance when combining the Inception architecture with residual connections.\footnote{See \citet{bianco2018benchmark} for an extensive benchmark analysis of popular convolutional neural network architectures.} Related to our setting for audio processing, \citet{li2019jasper} also uses residual convolutional neural networks for state of the art results on LibriSpeech dataset \cite{panayotov2015librispeech}. For the audio tower, we stack 3 replicas of the MFCCs and treat them as images. For each modality, a 1536-dimensional latent space representation is extracted. Despite using the same architecture for both encoders, their weights are not shared. Unless specified otherwise, the models are \textit{not} pretrained.

\begin{table*}
\setlength\tabcolsep{1.7pt}
    \centering
    \begin{tabular}{llcccccccccc}
    & & \multicolumn{5}{c}{\textbf{Caption to Image}} & \multicolumn{5}{c}{\textbf{Image to Caption}} \\\cmidrule(lr{0.1cm}){3-7}\cmidrule(l{0.1cm}){8-12}
    \multicolumn{2}{c}{Model} & R@1 & R@5 & R@10 & R@50 & R@100  & R@1 & R@5 & R@10 & R@50 & R@100\\\hline
    \multirow{4}{*}{Text} & \citealt{socher2014grounded} & - & - & .286 & - & - & - & - & .2r90 & - & - \\
    & \citealt{karpathy2014deep} & - & - & .425 & - & - & - & - & .440 & - & - \\
    & \citealt{harwath2015deep} & - & - & .490 & - & - & - & - & \textbf{.567} & - & - \\
    & \citealt{chrupala2017representations} & \textbf{.127} & \textbf{.364} & \textbf{.494} & - & - & - & - & - & - & - \\\Xhline{.5\arrayrulewidth}
    \multirow{7}{*}{Speech} & \citealt{harwath2015deep} & - & - & .179 & - & - & - & - & .243 & - & - \\
    & \citealt{chrupala2017representations} & .055 & 0.163 & .253 & - & - & - & - & - & - & - \\
    & \citealt{chrupala2018symbolic} & - & - & .296 & - & - & - & - & - & - & - \\\cmidrule(l{0.1cm}){2-12}
    & Ours (from scratch)         & .018 & .063 & .101 & .288 & .428 & .024 & .072 & .124 & .332 & .458 \\
    & Ours (warm-starting $g_{aud}$) & .041 & .138 & .211 & .467 & .613 & .550 & .166 & .241 & .522 & .654 \\
    & Ours (warm-starting $g_{vis}$) & .062 & .190 & .279 & .560 & .703 & .081 & .242 & .352 & .664 & .782 \\
    & Ours (warm-starting all) & \textbf{.139} & \textbf{.368} & \textbf{.495} & \textbf{.781} & \textbf{.875} & \textbf{.182} & \textbf{.435} & \textbf{.558} & \textbf{.842} & \textbf{.910}

    \end{tabular}
    \caption{Retrieval scores on the test set of FACC.}
    \label{tab:results_flickr}
\end{table*}

\paragraph{Optimization.} Models are trained using Adam \cite{kingma2014adam}, with an initial learning rate of 0.001 and an exponential decay of 0.999 every 1000 training steps, $\beta_1=0.9$, $\beta_2=0.999$ and $\epsilon=1\mathrm{e}{-8}$. We use a weight decay of $4\mathrm{e}{-5}$, and train on 32 GPUs until convergence. Unless specified otherwise, the optimization objective is minimizing the loss $\mathcal{L}_\text{MMS}$ (Eq. \ref{eq:loss}) with a margin term initially set to $\delta=0.001$ exponentially and increased by a factor of 1.002 every 1000 steps.

\subsection{Retrieval: Conceptual Spoken Captions}

Our primary aim with CSC is to use it for pretraining for later fine-tuning and evaluation on datasets with human speech instead of TTS. Nevertheless, we can compare different loss functions and different batch sizes on the CSC validation set to better understand the impact of these parameters.

We train models on CSC for 3 million steps, cropping/padding spoken captions to a duration of 3.5 seconds and using the loss functions $\mathcal{L}_\text{T}$ (Eq. \ref{eq:harwath_loss}) and $\mathcal{L}_\text{MMS}$ (Eq. \ref{eq:loss}). We find continuing improvements as batch size increases from 12 to 24 to 48. Furthermore, with the same batch size of 48, models optimized for minimizing $\mathcal{L}_\text{MMS}$ perform substantially better than those using $\mathcal{L}_\text{T}$, as summarized in Table \ref{tab:results_cac}. Of particular note is that R@1 scores for \lmms\ (batch size 48) are more than double those of \ltrip\ in both directions.

\subsection{Retrieval: Flickr Audio Caption Corpus}

Table \ref{tab:results_flickr} compares previous results on the FACC dataset with those obtained by variations of our model. As a pre-processing step, spoken captions are cropped/padded to a duration of 8 seconds. After pretraining the model in CSC, we explore all possible combinations of using or not the pretrained weights for each of the branches $g_{aud}$ and $g_{vis}$ as a warm-starting point, training until convergence on FACC. Warm-starting each of the branches in the dual-encoder leads to substantial improvements over the baseline, and combining both branches leads to the best overall performance. 

In particular, we improve R@10 for caption-to-image from the .296 obtained by \citet{chrupala2018symbolic} by 20\% absolute to .495, without using multitask training or pretraining $g_{vis}$ on ImageNet \cite{deng2009imagenet}. The multitask training approach of \citet{chrupala2018symbolic} is complementary to our improvements, so further gains might be obtained by combining these strategies. Furthermore, very deep, residual convolutional neural networks over characters have been shown to perform well for text-based tasks \cite{conneau-etal-2017-deep}. We expect that our strategy of using the same basic architecture across different input types (speech, text and image) can be fruitfully extended to that setting. A related observation: while our results exceed previous results reported on text/image retrieval settings for FACC, we expect that recent advances in text encoding could easily beat those reported numbers.

\begin{figure}
\centering
   \includegraphics[width=\linewidth]{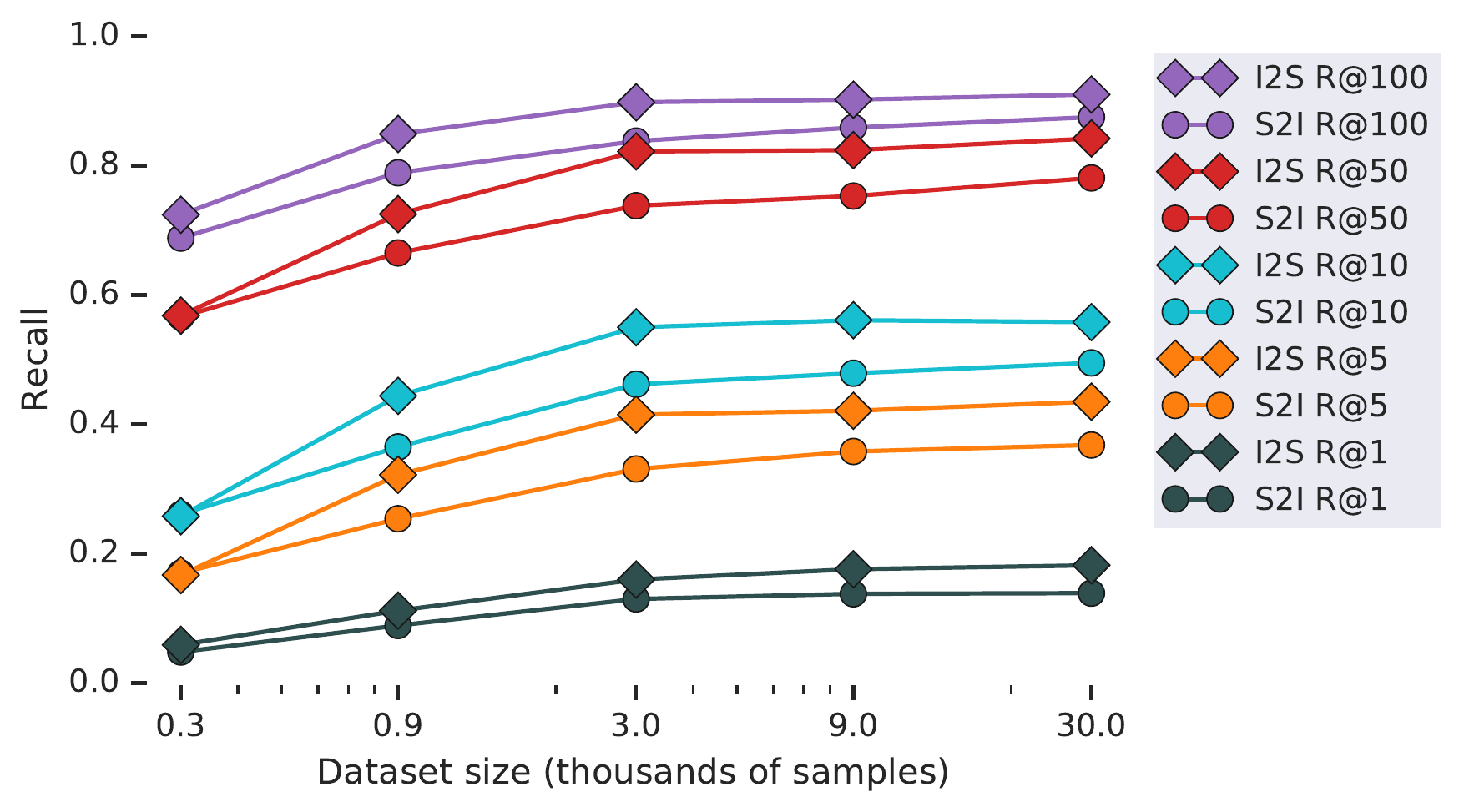}

  \caption{Ablations on low-data regime on FACC: chart shows recall scores for image-to-speech (I2S) and speech-to-image (S2I) retrieval, as a function of the amount of training data used for fine-tuning.}
\label{fig:lowdata}
\end{figure}

We also explore very low-data regimes using our pretrained model (see Fig. \ref{fig:lowdata}). Using small training subsets randomly drawn from FACC, we report performance as a function of how much data the model sees. With as little as 10\% of the original training data (3000 image/spoken caption pairs), the warm-started model performs competitively with a model trained on all training data.

\begin{figure*}[ht!]
  \centerfloat   
  \includegraphics[clip, width=1.0\linewidth]{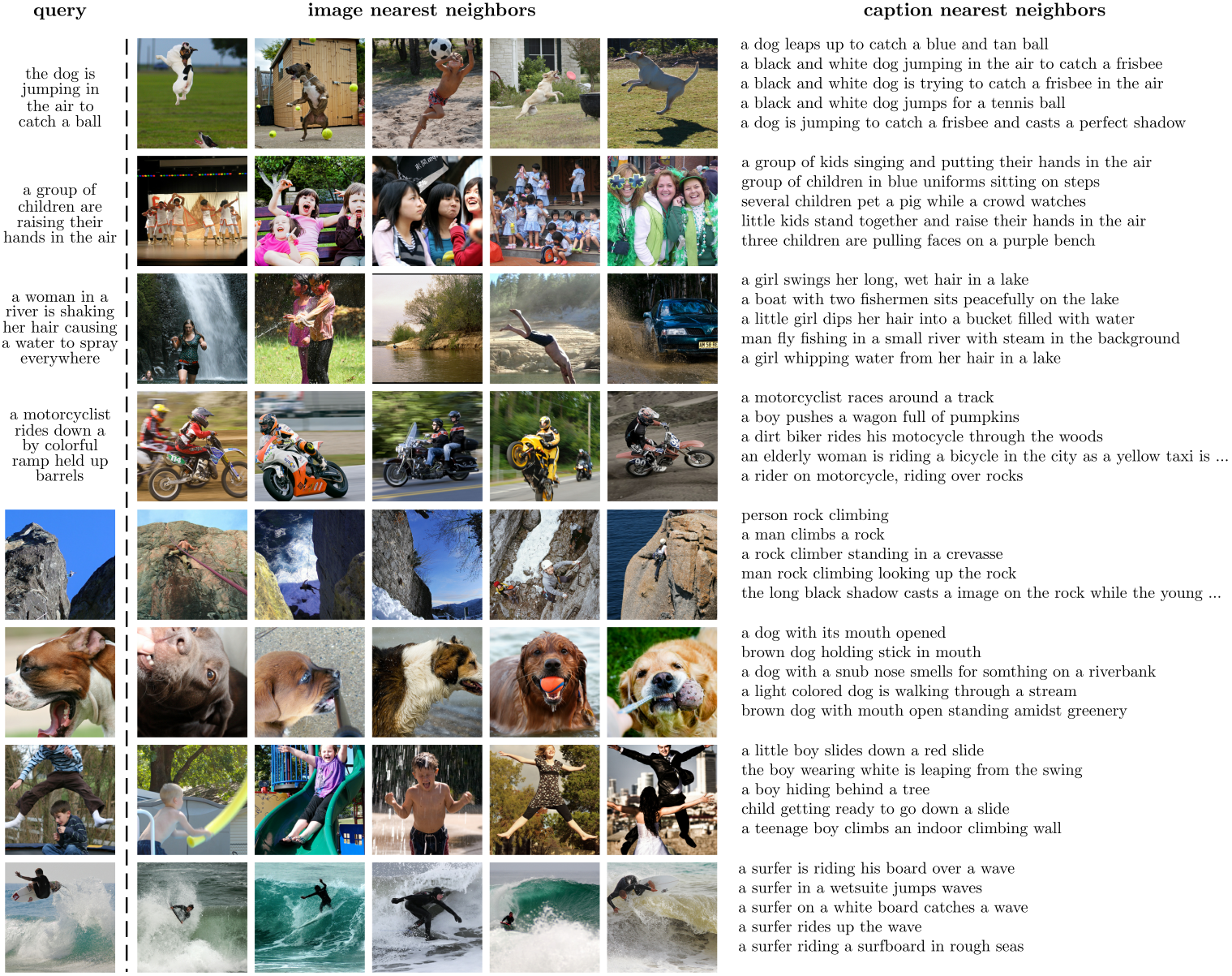}
  \caption{Nearest neighbors in the joint visual and acoustic latent space, best viewed with zoom: using 4 spoken captions and 4 images as queries, we extract from FACC's test set the closest 5 images and 5 spoken captions in the latent space for each of them. For simplicity, we show the text associated with each spoken caption.}
\label{fig:nn}
\end{figure*}
 
\paragraph{Qualitative evaluation.} Once a model is trained, any input (image or spoken caption) can be be used to query the corpus of images and spoken captions for nearest neighbors in the latent space. 
Figure \ref{fig:nn} shows some examples of retrieved nearest neighbors in FACC's test set. Given a spoken caption or an image we show the five nearest image neighbors and five nearest caption neighbors. From these, it is clear that the representations capture many semantically salient attributes of the inputs. The retrieved items correctly share many thematic elements and many are clearly good matches even though the particular image-caption pairs are not associated in the data. This serves to reinforce our observation that R@k evaluations using only the known paired items is likely to underestimate the actual performance of the models---which we show to be the case with human evaluations in Section \ref{sec:human_evals}.

Only some items are substantially incompatible: e.g. an image of a car for a caption about a woman in a river (they share water spraying), a picture of three adults for a caption about children raising their hands, and a caption about a boy climbing a wall for an image of children playing leapfrog). That said, many details are poor matches, such as the count of objects (one ball versus many), colors (brown dogs versus multicolored ones), people descriptions (elderly woman versus male dirt biker), object identification (e.g. a yellow pool noodle viewed as similar to slides), processes (jumping versus sliding) and perspective (man looking up versus viewed from behind and climbing). As such, there is clearly significant headroom for better, more fine-grained modeling of both captions and images. Additionally, cross-modal attention mechanisms \cite{xu2015show} and other explainability techniques \cite{ribeiro2016should} could help better inspect and understand a model's predictions.

Furthermore, as noted by \citet{chrupala2017representations}, text-based retrieval models often handle misspellings poorly. In contrast, speech-based models are unlikely to suffer from similar problems because they inherently must deal with variation in the expression of words and utterances. For instance, the caption \textit{``a dirt biker rides his \textbf{motocycle} through the woods"} (fourth row of Figure \ref{fig:nn}) is highly correlated with the correctly spelled sentences.

\subsection{Human evaluation}
\label{sec:human_evals}

We ran human evaluations to answer two questions: (1) how much does cropping limit model performance? and (2) how much do retrieval evaluations based only on positive associations underestimate model performance? Hints about both questions can be seen in the qualitative evaluation (Fig. \ref{fig:nn}).

To answer the first question, Table \ref{tab:human_evals_gt} shows the ratings for ground truth image/caption pairs in the FACC test set. The \textit{uncropped} row shows that overall the captions are high quality and do match the full images. However, human ratings on images \textit{cropped} at the center (which is what is provided to the models) show that there is considerable loss from cropping---only 62.5\% of cropped images are rated as good matches by all five raters. Inspection makes it clear why cropping hurts: for example an image of a snowboarder in the air next to another on a ski lift is cropped such that the snowboarder is missing, and thus a poor match to captions mentioning the snowboarder. This clearly indicates that standard cropping (which we follow) inherently limits model performance and that strategies that use the full image should be explored.

\begin{table}
\setlength\tabcolsep{2.8pt}
    \centering
    \begin{tabular}{l@{~~~}ccccc}
     & \multicolumn{5}{c}{``good" ratings (out of 5)}\\
     & 1+ & 2+ & 3+ & 4+ & 5 \\\hline
     Cropped & .949 & .918 & .874 & .800 & .625\\
     Uncropped & .995 & .994 & .989 & .971 & .891
    \end{tabular}
    \caption{Human evaluation results on ground truth pairs on the test set of FACC, using either center cropped (which the models receive) or uncropped versions of the images.}
    \label{tab:human_evals_gt}
\end{table}

Standard retrieval evaluations are blind to pairs that match but are not associated in the data. To address this and answer the second question posed above, we present the top-5 retrieved captions for each image and the top-5 retrieved images for each caption in FACC's test set to human raters. To increase speed and decrease costs, we show raters the original Flickr8k textual captions instead of the spoken ones. Each pair is judged by five raters as ``good" or not. This gives a soft measure of the compatibility of each pair based on fast binary judgments from each rater. For retrieval evaluations of a model, we compute recall based on the majority of human raters approving each image-caption pair: R@1 is the percentage of top-1 results and R@5 the percentage of top-5 results that are evaluated as a match by at least 3 of the 5 raters.

Table \ref{tab:human_evals} shows these metrics computed on retrieval outputs from two settings: FACC training from scratch and FACC fine-tuning after CSC pretraining. It also shows the corresponding automatic evaluations from Table \ref{tab:results_flickr} for easy comparison. These results make it clear that evaluation based only on positive associations is too rigid: speech-to-image retrieval based on human evaluations shows that a good matching item is returned in 52.2\% of cases rather than just the 36.8\% indicated by strict corpus matches. For image-to-speech retrieval the 55.8\% strict measure goes up to 63.8\%. That said, the results also show that the strict measure is nevertheless a useful indicator for comparing relative model performance: the model pretrained on CSC beats the corresponding one trained on FACC from scratch, on both human and automatic evaluations.

\begin{table}
\setlength\tabcolsep{3.2pt}
    \centering
    \begin{tabular}{lc@{~~~}ccccccc}
     & & \multicolumn{2}{c}{S2I} & \multicolumn{2}{c}{I2S}\\\cmidrule(l{-0.01cm}r){3-4}\cmidrule(l{0.1cm}r){5-6}
    Eval & Pretrain  & R@1 & R@5 & R@1  & R@5 \\\hline
    Auto  &        & .018 & .063 & .024 & .072 \\
    Auto  & \cmark & .139 & .368 & .182 & .558 \\\hline
    Humans  &        & .056 & .154 & .070 & .196 \\
    Humans  & \cmark & .229 & .522 & .306 & .638 \\
    \end{tabular}
    \caption{Comparison of human rater scores (majority agreement) versus using only corpus-known pairs on all metrics for speech-to-image (S2I) and image-to-speech (I2S) retrieval. Rows with \textit{Auto} evaluation correspond to \textit{Ours (from scratch)} and \textit{Ours (warm-starting all)} scores in Table \ref{tab:results_flickr}.}
    \label{tab:human_evals}
\end{table}

\section{Conclusion}

Large-scale datasets are essential for training deep networks from scratch. In this paper, we present a scalable method for generating an audio caption dataset taking advantage of TTS systems to create millions of data pairs. Using the MMS loss, we demonstrate that pretraining on this dataset greatly improves performance on a human-generated audio caption dataset. As TTS models continue to improve and be developed for more languages, this data augmentation strategy will only become more robust and helpful over time. Finally, using human evaluations, we show evidence that corpus-based retrieval scores underestimate actual performance.

This present work is focused on the here and now since captions describe a snapshot in time and focus on the visual entities and events involved in them. We thus have little hope to learn representations for words like \textit{visit}, \textit{career} and \textit{justice}, for example. Videos can help with process oriented words like \textit{visit} and could get significant components of words like \textit{career} (such as the visual contexts, but not the overall path with intermediate goals involved in careers). They are likely to be hopeless for abstract words like \textit{justice}. To address problems of this sort, there are likely many opportunities to combine ideas from unsupervised term discovery \cite{kamper.jansen.ea:unsupervised,bansal:etal:2017} with audiovisual word learning \cite{harwath2018jointly} and models of visual grounding that have been applied to text \cite{kiros-etal-2018-illustrative}.
Being able to learn effective representations from raw audio associated with images could provide new possibilities for work that learns from videos and text (transcribed speech) \cite{chen-etal-2018-temporally}, and in particular open up such techniques to new languages and domains.

\section*{Acknowledgements}
The authors would like to thank Radu Soricut, Austin Waters, Alex Ku and Jeffrey Ling for the helpful comments that assisted the development of this work.

\bibliography{main}
\bibliographystyle{acl_natbib_nourl}

\end{document}